\documentclass{article}
\usepackage{spconf,amsmath,graphicx}
\usepackage{multirow,threeparttable, booktabs, makecell, pifont}
\usepackage{tabularx}
\usepackage{amssymb}
\usepackage{lipsum}
\usepackage{tabularx}

\usepackage{subfig}
\usepackage{float}

\usepackage{bbding}
\usepackage{algorithmicx}
\usepackage{algorithm}
\usepackage{algpseudocode}
\usepackage{amsmath} % for \boldsymbol macro
\usepackage{color}
\usepackage{graphicx}
\usepackage{wrapfig}
\title{Enhancing Generalization in Medical Visual Question Answering Tasks via Gradient-Guided Model Perturbation}
\name{
\begin{tabular}{c}
\it Gang Liu$^1$, Hongyang Li$^{1*}$, Zerui He$^1$, Shenjun Zhong$^{2*}$\thanks{* Corresponding author. 
This work was supported by the Ministry of Education Humanities and Social Science Research Planning Fund Project under grant number 23YJAZH079 and Natural Science Foundation of Heilongjiang Province under grant number LH2021F015.
}
\end{tabular}
}
\address{
    $^1$College of Computer Science and Technology, Harbin Engineering University, China,\\
    $^2$Monash Biomedical Imaging, Monash University, Australia}

\begin{document}
%\ninept
%
\maketitle
\begin{abstract}
\vspace{-2pt}
Leveraging pre-trained visual language models has become a widely adopted approach for improving performance in downstream visual question answering (VQA) applications. However, in the specialized field of medical VQA, the scarcity of available data poses a significant barrier to achieving reliable model generalization. Numerous methods have been proposed to enhance model generalization, addressing the issue from data-centric and model-centric perspectives. Data augmentation techniques are commonly employed to enrich the dataset, while various regularization approaches aim to prevent model overfitting, especially when training on limited data samples. In this paper, we introduce a method that incorporates gradient-guided parameter perturbations to the visual encoder of the multimodality model during both pre-training and fine-tuning phases, to improve model generalization for downstream medical VQA tasks. The small perturbation is adaptively generated by aligning with the direction of the moving average gradient in the optimization landscape, which is opposite to the directions of the optimizer's historical updates. It is subsequently injected into the model’s visual encoder. The results show that, even with a significantly smaller pre-training image caption dataset, our approach achieves competitive outcomes on both VQA-RAD and SLAKE datasets.
\end{abstract}
\vspace{-6pt}
\section{INTRODUCTION AND RELATED WORK}
\label{sec:intro}
\vspace{-3pt}

Visual Question Answering (VQA) is a task that aims to predict answers to questions based on the content within images. In the specialized field of Medical VQA, a significant challenge is the restricted size of publicly accessible datasets, often due to privacy considerations. A widely used strategy to address this limitation involves a pre-training and finetuning approach, where medical image caption datasets are typically used to learn a vision-language model which is subsequently transferred to medical VQA tasks, as demonstrated by the methods like M3AE\cite{M3AE} and M2I2\cite{c13}.
 
Within a pre-training and finetuning framework, to further improve the model generalization, data augmentation as a popular data-centric technique adopted by existing works \cite{c3,c6} that expands the training dataset by applying various transformations (e.g. rotation and cropping of images), without the need for collecting additional data. To prevent overfitting, auxiliary regularization techniques are also commonly used in the training process, such as DropOut\cite{dropout}, L2 regularization \cite{L2}, and noise injection\cite{Parametric}. 
Some works that add noise or perturbations to the model’s weights are mainly designed to improve the model's robustness against adversarial attacks \cite{Parametric}. The most relevant work is FlipOut \cite{Flipout} which introduces Gaussian perturbations to the model weights, and demonstrates superior regularization capabilities compared to the dropout methods.

In the context of a pre-training and fine-tuning framework, we further explore the options of introducing dynamic perturbations to model weights during training, to improve the model generalization on downstream medical VQA tasks. Our proposed method generates gradient-guided perturbations and integrates them into the visual encoder of the multi-modality model tailored for medical VQA tasks, in both pre-training and finetuning phases. The method is validated VQA-RAD\cite{c17} and SLAKE\cite{c10} datasets and achieves an overall accuracy of 78.94\% and 85.2\% respectively. The results show that the proposed method demonstrates competitive performance with significantly fewer data samples in the pre-training stage.

%\\\texttt{papers@2021.ieeeicassp.org}.
\vspace{-17pt}
\section{METHOD}
\label{sec:METHOD}
\vspace{-8pt}

\begin{figure*}[htbp]
    \vspace{-6pt}
    \centering
    \includegraphics[width=0.8\textwidth]{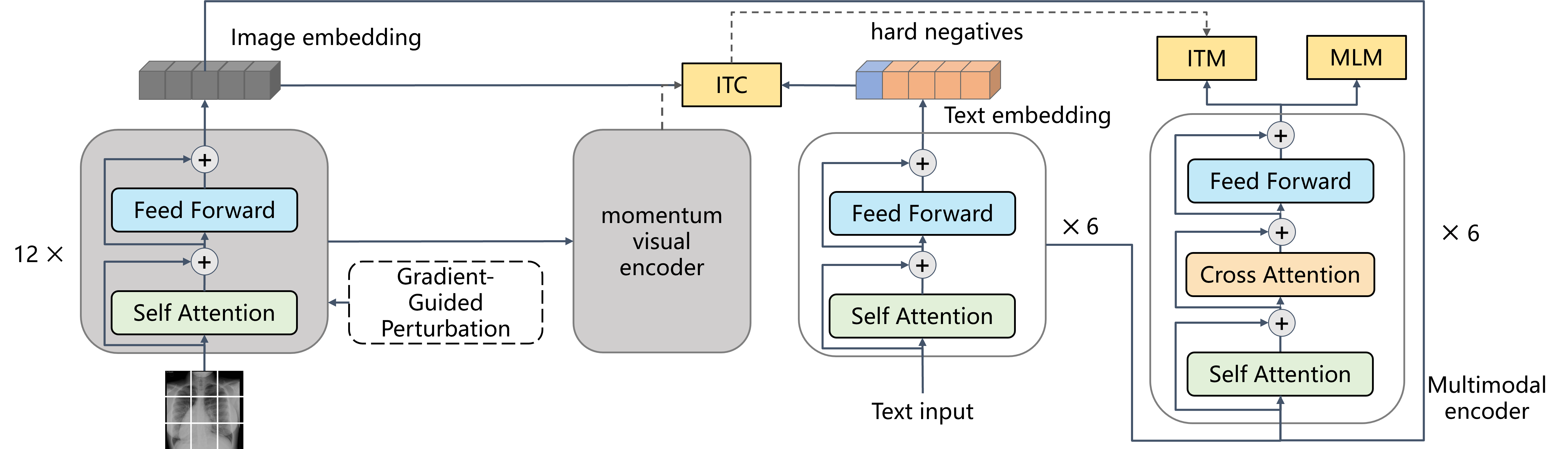}
    \vspace{-6pt}
    \caption{Pre-training model architecture and objectives of the proposed method. }
    \label{fig1}
    \vspace{-15pt}
\end{figure*}

In this section, we introduce the method that integrates gradient-guided perturbations into the pre-training and finetuning framework for VQA tasks. These adaptive model perturbations are applied only to the visual encoder of the model during each training iteration in both the pre-training and fine-tuning stages. Besides, we employ multiple self-supervised objectives for the vision-language pre-training (VLP) phase on an image caption dataset, and the pre-trained model is then fine-tuned for downstream VQA tasks. Further details will be described in the following sections.

\vspace{-7pt}
\subsection{Gradient-Guided Perturbation Generation}%======================================
\vspace{-2pt}

The proposed technique to construct model perturbations leverages the optimization momentum maintained by the Adam optimizer\cite{c18}, where the momentum is the exponentially moving average of the recent gradients that indicate the directions of the recent model updates. For a given an input image $x$, the averaged gradient $\nabla_{t}$ at time $t$, can be computed by the equation below:
\[
\nabla_{t} = \beta\nabla_{t-1}+(1-\beta)\cdot \nabla \mathcal L(\theta_{t},\mathcal{M}(x),y_{true})
\]
where $\mathcal{M}(x)$ denotes the output yielded by the model, and $y_{true}$ is the ground truth of the training sample; $\theta_{t}$ is the current model weights, and $\nabla \mathcal{L}$ is the gradient calculated based on the loss at the current iteration.; Besides, $\beta$ serves as the hyper-parameter that controls the decay rates of these moving averages.  The moving-averaged gradients $\nabla_{t}$, evolve through the entire training process.     

In our method, we specifically design the perturbations for the model weights of the visual encoder. The perturbations are formed as a minor offset along the direction of the moving averaged gradients of past model updates. In other words, the perturbations introduced to the visual encoder are against the direction of the most recent updates made by the optimizer in the optimization landscape. The gradient-guided perturbation, denoted as $r_{t}$ at a given time $t$ can obtained by the following equation:
\[
r_{t} = \delta \cdot \frac{\left\|\theta_{t} \right\|}{\left\| \nabla_{t} \right\|} \cdot \nabla_{t}
\]
where  $\theta_{t}$ and $\nabla_{t}$ are the model weights and the moving averaged gradients of the visual encoder at the current iteration, respectively. The ratio between the L2 norm of the model weights $\theta_{t}$ and the gradient $\nabla_{t}$ serves as an indicator for the scale of the gradients relative to the weights. This ratio is used to adjust the magnitude of perturbations added to the model in an adaptive manner. Specifically, a smaller ratio suggests the current iteration is still in the “steep” regions of the loss landscape, while a larger ratio indicates the optimization is at a “flat” region where a larger perturbation is applied. Besides, $\delta$ acts as the perturbation coefficient,  another hyper-parameter to control the magnitude of the perturbation.  

As the gradients and model parameters update, an adaptive perturbation, $r_{i}$ is added to the visual encoder in each iteration. The objective here is to modify the model weights in the opposite direction of the prior optimizations in the loss landscape. To minimize the risk of gradient explosion, we further apply clipping on the perturbed model weights as per the following equation:
\[
\theta^{'}_{t} = \theta_{t-1}+ \mathcal Clip( r_{i}, -\epsilon\cdot|\theta_{t-1}|, \epsilon\cdot|\theta_{t-1}|)
\]
where $\mathcal Clip()$ is a clipping function, ensuring that the perturbation remains within a  range that is governed by the clipping margin,  $\epsilon\cdot|\theta_{t-1}|$. $\epsilon$ is an additional hyper-parameter to control the scale of the margin; in our case, it takes a small value, such as 0.001. $\theta_{t-1}$ refers to the recently updated model parameters, i.e. those following the last model update. In our method, the addition of perturbations is regarded as a regularization technique for mitigating overfitting.

The associated pseudo-code for adding perturbations to the training process is presented in Algorithm 1. During each iteration, we obtain the first-order moment from the AdamW optimizer to generate the adaptive weight perturbation. The losses are calculated using the perturbed model, rather than the original weights. Subsequently, the gradients are updated to the unperturbed weights that are stored prior to each iteration.

%======================================
\setlength{\textfloatsep}{16pt}% Remove \textfloatsep
\begin{algorithm}[!h]
	\caption{
		Pseudo code of perturbation in pre-training
	}
	\label{alg:freemat}
	\begin{algorithmic}[1]
    	\Require Training samples $\mathcal{D}=\{({\boldsymbol x}_{img}, {\boldsymbol x}_{text}, {\boldsymbol y})\}$
    	\For{epoch $= 1 \ldots N$}
        	\For{minibatch $B\subset D$}

		\State Take a snapshot of $\theta_{t-1}$
            \State Compute the first-order gradient moment $\nabla_{t}$
            \State Generate the  perturbation $r_{t}$
            \State Get perturbed model ${\boldsymbol \theta}^{'}_{t}$, given $\boldsymbol \theta_{t-1}, r_{t}$
            \State Compute losses with perturbed model $\boldsymbol \theta^{’}_{t}$
		\State Update model ${\boldsymbol \theta_{t-1}} \rightarrow {\boldsymbol \theta_{t}}$
     	\EndFor
        
    	\EndFor
	\end{algorithmic}
\end{algorithm}
\vspace{-6pt}

In our work, we add perturbations in both the pre-training and finetuning phases and conduct validation on the effectiveness of this technique. Importantly, we narrow the focus of our investigation by only applying the perturbations to the visual encoder of the model.

\vspace{-6pt}
\subsection{Pre-training on Image-Text Pairs}
\label{ssec:subhead}
%\vspace{-3pt}

\subsubsection{\textbf{Network Architecture}}
\label{sssec:subsubhead}

As shown in Fig.1, the overall network architecture of the multi-modality model, designed for pre-training on image caption datasets, contains three main components: a 12-layer transformer-based visual encoder, a 6-layer transformer-based text encoder, and a multi-modality encoder to fuse visual and text information. The visual encoder is initialized with the pre-trained weights from ViT\cite{ViT}, while the text encoder is initialized with the first 6 layers of a pre-trained BERT model. In line with the approach proposed in the MoCo\cite{MOCO}, separate visual and text momentum-updated models are maintained during training. They serve to construct more stable visual and text representations for contrastive training objectives.
 
During the pre-training phase, an image caption dataset is used to learn both image and text feature representations, as well as their iterations. Input image patches are fed into the visual encoder with a special token [CLS].  The embedding corresponding to the [CLS] token is treated as the image embedding and forwarded to the multi-modality encoder. Similarly, the embeddings of the [CLS] token in the text inputs encode the semantics of image captions. 
In each training iteration, gradient-guided perturbations are added to the visual encoder as we described in the previous section. The perturbed visual encoder is then employed to encode the input image. Subsequently, the encoded lower-dimensional (256-d) representations are used to calculate the following three pre-training objectives.

%%haimeigai
\vspace{-5pt}
\subsubsection{\textbf{Pre-training Objectives}}
\label{sssec:subsubhead}

\textbf{Image-Text Contrastive Learning (ITC).}
We use the ITC training strategy, initially proposed by ALBEF\cite{albef}, as one of the pre-training objectives. This method aims to bring an image and its associated caption closer in the feature space, while simultaneously distancing them from irrelevant or negative captions. Similar to the MoCo approach, we maintain a buffer of image-text embedding pairs, where the unassociated image-text combinations are used as negative samples to compute the contrastive loss. In this work,

\textbf{Image-Text Matching (ITM).}
 Another training objective used in the pre-training phase is ITM loss. It is formed as a binary classification task where the input is the joint representation of the image embedding and text embedding. The task is to predict whether a given image and its corresponding text (or caption) are a match, with the negative pairs being sampled from the current training batch. The ITM loss for a   specific image-text pair can be formulated as:
\[
\mathcal{L}_{ITM}= \mathcal{L}_{CE}(p_{itm}(I',T),y_{itm}) 
\]
$p_{itm}$ represents the prediction results of image text pairs, and $y_{itm}$ is ground truth, $\mathcal{L}_{CE}$ represents the calculation process of the cross entropy loss function.

\textbf{Masked Language Modeling (MLM).}
Similar to the BERT approach, we employ an MLM task aimed at predicting masked tokens in an input text based on their contextual information, which includes the associated input image and the surrounding text tokens in the caption. In our implementation, 15\% of tokens in the image caption are randomly masked and replaced with a special token, [MASK]. Images and the masked caption are then processed through the multi-modal encoder to output the joint representation. The joint representation is subsequently passed to the decoder to predict the probabilities of the masked tokens. In our method, the MLM loss is computed using the cross-entropy loss between the predicted and actual tokens.   
%%daozheli
\vspace{-6pt}
\subsection{Finetuning on Medical VQA}

\vspace{-15pt}

\begin{figure}[htbp]
    \centering
    \includegraphics[width=\linewidth]{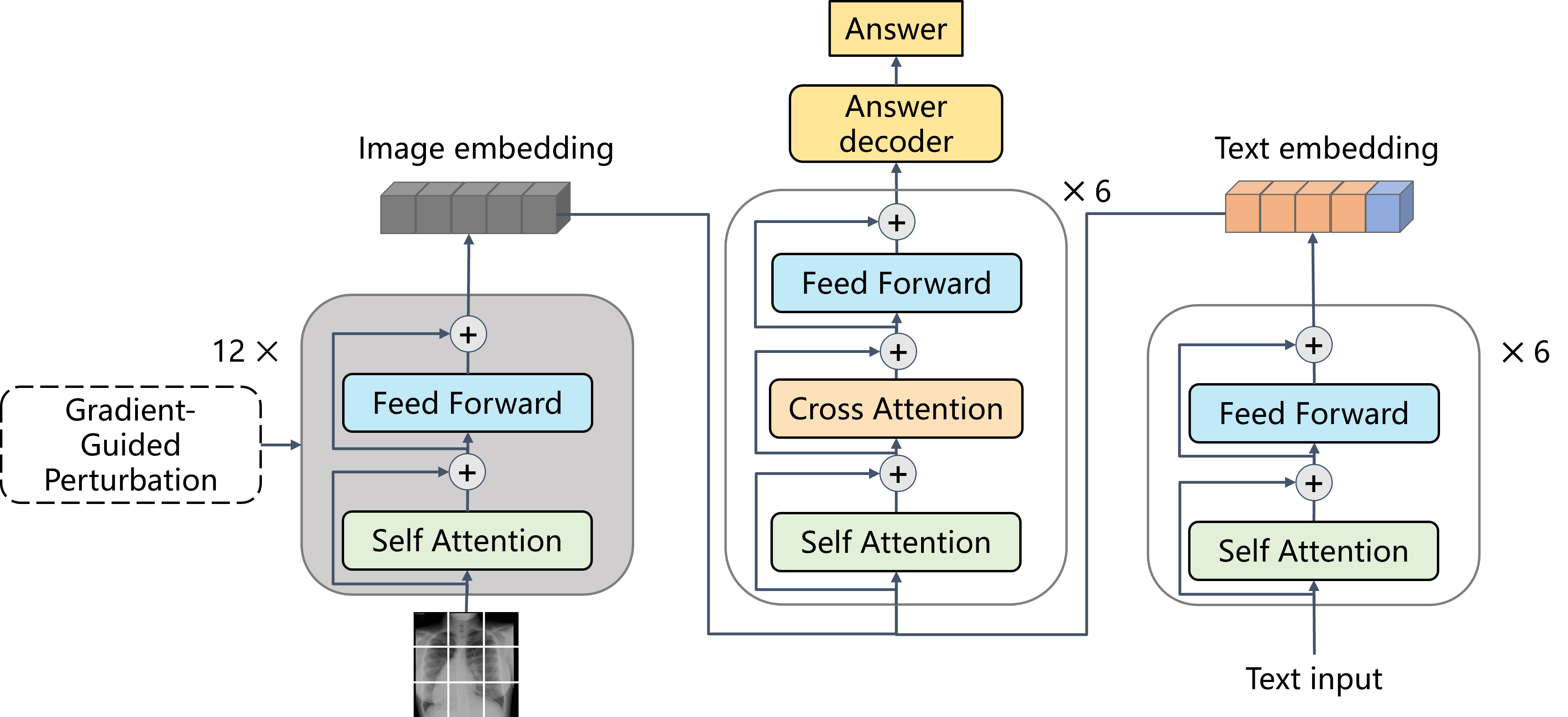}
    \vspace{-12pt}
    \caption{Finetuning model structure.}
    \label{fig2}
    \vspace{-10pt}
\end{figure}

During the finetuning phase for the downstream VQA task, an answer decoder is added, as shown in Fig 2, which is a BERT-style decoder consisting of a 6-layer transformer,  initialized with the pre-trained weights from BERT, and a prediction head to generate answer tokens generatively. Similar to the pre-training phase, we also introduce model perturbations in the fine-tuning process. The gradients for these perturbations are computed on the fly, from the conditional language-modeling loss.

\vspace{-15pt}
\section{EXPERIMENT}
\label{sec:EXPERIMENT}
\vspace{-6pt}

\vspace{-3pt}
\subsection{Datasets}
\vspace{-1pt}
%数据集介绍
We used the ROCO\cite{c9} dataset to pre-train the model.  This dataset contains about 81,000 radiology images from various medical imaging modalities, sourced from the open-access PubMed Central database. The images in the ROCO dataset have corresponding captions of various lengths that describe the content of the images. 

In the downstream medical VQA tasks, two datasets, VQA-RAD and SLAKE are used. VQA-RAD\cite{c17} is a specialized radiology-based VQA dataset containing 3515 questions and 517 answers about 315 images. These question-answer (QA) pairs are prepared and validated by medical professionals. We follow the official dataset partition for our evaluation measurements, with the training set that consists of 3064 question-answer pairs, and the test set containing 451 pairs. SLAKE \cite{c10} is another publicly available medical VQA dataset, which includes 14,000 QA pairs across 642 images. Both datasets offer two question types, open-ended and closed-ended. Closed-ended answers are limited to yes and no, while open-type answers can be any free-form text.

\vspace{-9pt}
\subsection{Result}
\vspace{-3pt}

We use the AdamW\cite{c18} optimizer for our pre-training tasks, with a learning rate of 3e-4, a batch size of 32, and a 5\% warm-up ratio to pre-train on the image caption dataset for 30 epochs. For the downstream medical VQA tasks, we fine-tune the model for an additional 40 epochs, using the same optimizer with a smaller learning rate of 3e-5.

We transferred the weights pre-trained on the image caption dataset and evaluated the performance on two downstream medical VQA datasets. In Table 1, it shows the performance comparison of our method’s prediction accuracy against existing methods on the VQA-RAD and SLAKE datasets.  The proposed method outperforms the previous state-of-the-art (SOTA) methods on VQA-RAD, with an overall accuracy of 78.94\% (70.39\% on open-ended questions and 84.56\% on closed-ended questions). Notably, it achieves a 2.14\% higher accuracy on VQA-RAD, compared to M2I2 which is also pre-trained on an image-caption dataset of similar size. Besides, it is worthwhile highlighting that the proposed method beats M3AE and PMC-VQA, even though they are trained on much larger datasets.

On the SLAKE dataset, our method achieves an overall accuracy of 85.2\%. The best result 88.0\%, is from MedVInT-TE which is pre-trained on the PMC-VQA dataset that is a refined and better quality dataset compared to ROCO. While our method outperformed MedVInT-TE on closed-ended questions by 2.2\%, it lags behind the M2I2 approach. For open-ended questions, our approach yields an accuracy of 82.17\%, which is lower than the SOTA result of 88.0\% set by MedVInT-TE, but surpassing other methods in this category.

\begin{table}[!htb]

\centering
\tabcolsep=0.13cm
\caption{Comparisons with the state-of-the-art methods on the VQA-RAD and SLAKE test set.}

\vspace{-5pt}
\resizebox{\linewidth}{!}{
\begin{tabular}{cccccccc}

\hline
\multirow{2}{*}{\textbf{Method}} & \multirow{2}{*}{\makecell[c]{\textbf{Pre-Training Data}}} & \multicolumn{3}{c}{\textbf{VQA-RAD}} & \multicolumn{3}{c}{\textbf{SLAKE}} \\
& & \multicolumn{1}{c}{Open} & \multicolumn{1}{c}{Closed} & \multicolumn{1}{c}{Overall} & \multicolumn{1}{c}{Open} & \multicolumn{1}{c}{Closed} & \multicolumn{1}{c}{Overall} \\ \hline

MEVF\cite{c8} & - & 43.9 & 75.1 & 62.6 & - & - & - \\

CPRD\cite{c7} & - & 61.1 & 80.4 & 72.7 & 81.2 & 83.4 & 82.1 \\ 

AMAM\cite{c2} & - & 63.8 & 80.3 & 73.3 & - & - & - \\

M2I2\cite{c13} & 91k & 66.5 & 83.5 & 76.8 & 74.7 & 91.1 & 81.2 \\

M3AE\cite{M3AE} & 401k & 67.2 & 83.5 & 77.0 & 80.3 & 87.8 & 83.2 \\

PMC-VQA\cite{pmcvqa} & 177k & 69.3 & 84.2 & 78.2 & \hspace{5pt}\textbf{88.2}\hspace{5pt} & \hspace{5pt}87.7\hspace{5pt} & \hspace{5pt}\textbf{88.0}\hspace{5pt} \\

ours & 81k & \hspace{5pt}\textbf{70.39}\hspace{5pt} & \hspace{5pt}\textbf{84.56}\hspace{5pt} & \hspace{5pt}\textbf{78.94}\hspace{5pt} & 82.17 & \hspace{5pt}\textbf{89.9}\hspace{5pt} & 85.2 \\ \hline

\end{tabular}
}

\label{tab1} 
\vspace{-5pt}

\end{table}

\textbf{Ablation Study.}To study the effectiveness of the gradient-guided perturbation on both the pre-training and finetuning stages, we conduct ablation experiments that introduce these perturbations in (i) only the pre-training (PT) stage, (ii) only the finetuning (FT) stage; or (iii) both. Besides, we explore the contribution of adaptive perturbation magnitude (APM) in our work. As shown in Table 2, the baseline performance is obtained from the models trained without any perturbations. When we incorporate perturbations solely during the pre-training stage (i.e. no perturbations are added to the model for downstream VQA tasks), there is a marginal improvement of  0.22\% for VQA-RAD and 0.57\% for SLAKE compared to the baseline. In contrast, applying perturbations only during the downstream VQA tasks results in accuracies of 78.05\% and 83.88\% on VQA-RAD and SLAKE respectively. Furthermore, we investigate the effectiveness of using adaptive magnitudes for the applied perturbations. In this experiment,  a fixed weighting value of 0.05 is used, instead of an adaptive value that is calculated dynamically based on the L2 norm of the model and the gradients. Without this adaptiveness, the model performs less optimally, achieving 78.94\% on VQA-RAD and 85.2\% on SLAKE, which is lower than the results of the proposed method.

\begin{table}[!htb]
\vspace{-5pt}
\centering
\caption{The ablation study validated on VQA-RAD and SLAKE dataset. PT and FT mean perturbing the model in the pre-training and fine-tuning phases respectively, and APM means adaptive perturbation magnitude.}
\resizebox{0.6\linewidth}{!}{
\begin{tabular}{lcccccc}
\toprule

PT & FT & APM & VQA-RAD & SLAKE   \\
  \midrule 
  & &  & 77.16 & 82.09 \\
 \Checkmark &  & \Checkmark & 77.38  & 82.66 \\
   & \Checkmark & \Checkmark  & 78.05  & 83.88 \\
 \Checkmark & \Checkmark & & 77.83 & 82.94 \\
 \Checkmark & \Checkmark & \Checkmark & \textbf{78.94} & \textbf{85.2}  \\
\bottomrule
\end{tabular}
}
\label{tab2}
\vspace{-5pt}
\end{table}

\vspace{-10pt}
\begin{figure}[htbp]
    \centering
    \includegraphics[width=0.8\linewidth]{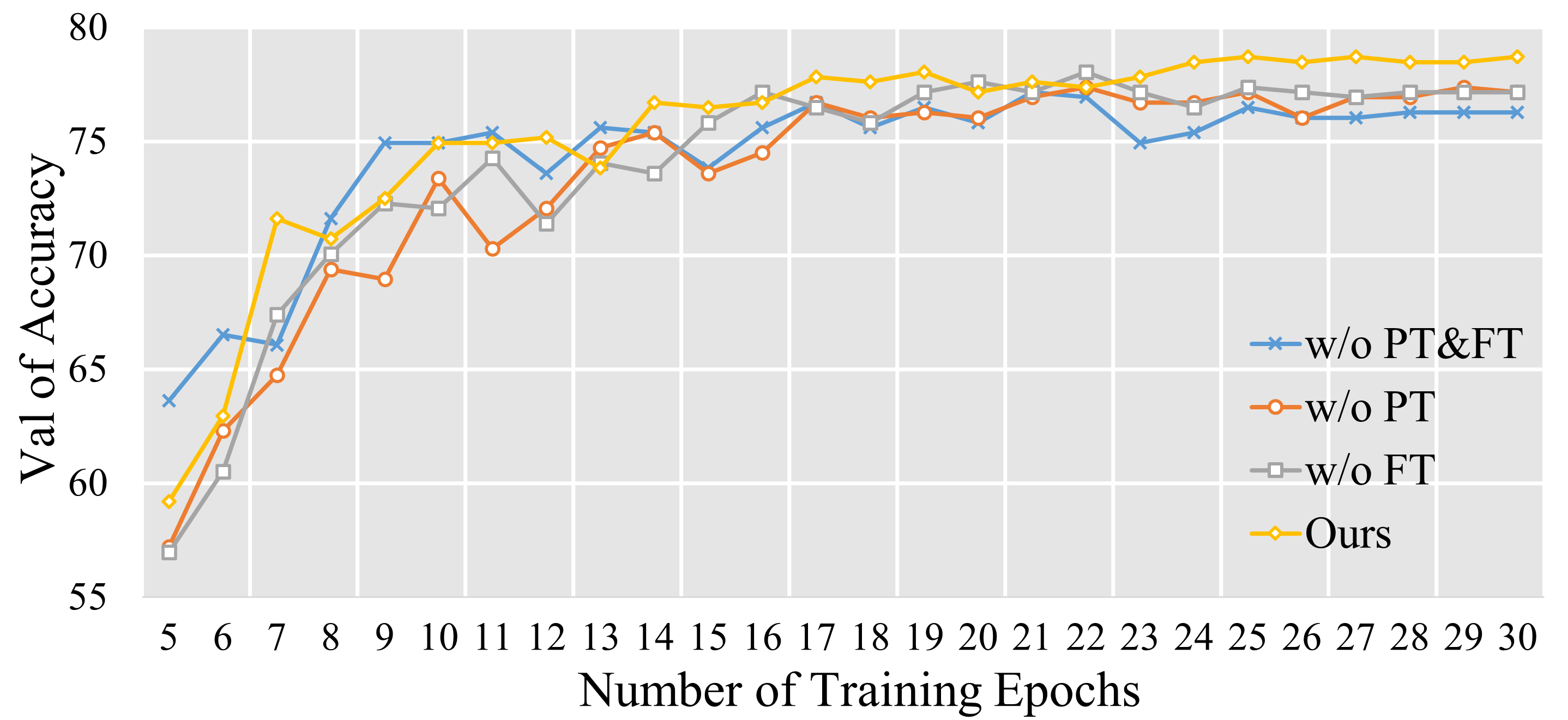}
    \vspace{-3pt}
    \caption{Validation accuracies on VQA-RAD dataset during downstream VQA training, with different components enabled.}
    \label{fig3}
    \vspace{-10pt}
\end{figure}

The method is designed as a regularization technique, therefore, we visualize the validation accuracy during training in Fig 3, to demonstrate its effectiveness. The plot shows that introducing gradient-guided perturbations in both pre-training and downstream tasks leads to a higher and more stable validation accuracy growth. This is especially notable when the performance gap becomes consistent after 24 epochs in the downstream VQA tasks.

\vspace{-10pt}
\section{CONCLUSION}
\label{sec:typestyle}
\vspace{-7pt}

In this paper, we propose a novel regularization technique for vision-language pre-training, that introduce adaptive perturbations to the visual encoder of the multi-modality model to enhance model generalization. The method can be applied to both pre-training and downstream medical VQA tasks. The results show that the method improves the performance on downstreammedical VQA tasks, while requiring less data for pre-training. Importantly, our approach is model-agnostic, making it potentially useful in other uni-modal or multi-modal applications.

% \section{ACKNOWLEDGMENTS}
% \label{sec:page}

\vfill\pagebreak

\bibliographystyle{IEEEbib}
\bibliography{strings,refs}

\end{document}